%% file: main.tex
\documentclass[conference]{IEEEtran}
\IEEEoverridecommandlockouts

\usepackage{cite}
\usepackage{amsmath,amssymb,amsfonts}
\usepackage{algorithmic}
\usepackage{graphicx}
\usepackage{subfig}
\usepackage{float}
\usepackage{textcomp}
\usepackage{xcolor}
\usepackage{xspace}
\usepackage{array,booktabs} 
\usepackage{graphbox} 

\usepackage[draft]{todonotes} 
\def\BibTeX{{\rm B\kern-.05em{\sc i\kern-.025em b}\kern-.08em
    T\kern-.1667em\lower.7ex\hbox{E}\kern-.125emX}}
\usepackage{hyperref}

\newcommand{\vago}{\texttt{VAGO}\xspace}
\newcommand{\fake}{\texttt{FAKE-CLF}\xspace}

\begin{document}

\title{Combining vagueness detection with deep learning to identify fake news}

\author{\vspace{0.05in}
\IEEEauthorblockN{Paul Gu\'elorget\IEEEauthorrefmark{1}, Benjamin Icard\IEEEauthorrefmark{2}, Guillaume Gadek\IEEEauthorrefmark{1}, Souhir Gahbiche\IEEEauthorrefmark{1},\\ Sylvain Gatepaille\IEEEauthorrefmark{1}, Ghislain Atemezing\IEEEauthorrefmark{3} and Paul \'Egr\'e\IEEEauthorrefmark{2}}

\vspace{0.2in}
\IEEEauthorblockA{\IEEEauthorrefmark{1}\textit{Advanced Information Processing}, Airbus, France}

\vspace{0.03in}
\IEEEauthorblockA{\IEEEauthorrefmark{2}\textit{Institut Jean-Nicod}, CNRS, ENS, EHESS, PSL University, France}

\vspace{0.03in}
\IEEEauthorblockA{\IEEEauthorrefmark{3}Mondeca, Paris, France}
}

\maketitle

\begin{abstract} %
 In this paper, we
combine two independent detection methods for identifying fake news: the algorithm \vago uses semantic rules combined with NLP techniques to measure vagueness and subjectivity in texts, while the classifier \fake relies on Convolutional Neural Network classification and supervised deep learning to classify texts as biased or legitimate. We compare the results of the two methods on four corpora. We find a positive correlation between the vagueness and subjectivity measures obtained by \vago, and the classification of text as biased by \fake. The comparison yields mutual benefits: \vago helps explain the results of \fake. Conversely \fake helps us corroborate and expand \vago's database. The use of two complementary techniques (rule-based vs data-driven) proves a fruitful approach for the challenging problem of identifying fake news.

\end{abstract}

\begin{IEEEkeywords}
vagueness, subjectivity, bias, fake news, deep learning, expert system, explainable AI

\end{IEEEkeywords}

\section{Introduction}
    \label{sec:intro}

\input{texts/1intro}

\section{Vagueness detection: \texttt{VAGO}}
    \label{sec:vago}
    \input{texts/2vago}

\section{Fake news classifier}
    \label{sec:deep}

\input{texts/3deep}

\section{Comparison and combination}
    \label{sec:correlation}
    \input{texts/4comparison}

\section{Discussion}
    \label{sec:discussion}
    \input{texts/5discussion}

\section{Conclusion}
    \label{sec:conclusion}
    \input{texts/6conclusion}

\section*{Acknowledgments}
{ This work was funded by the project ``Disinformation Identification in Evolving Knowledge Bases'' (DIEKB) under grant n° DGA01D19018444 (Mondeca, CNRS, Airbus). PE and BI also acknowledge grants ANR-19-CE28-0004-01 (Probasem), ANR-19-CE28-0019-01 (Ambisense), and ANR-17-EURE-0017 (FrontCog) for support. We thank three anonymous reviewers and Denis Bonnay for their helpful comments.}

\section*{Declaration of contribution}

{The three teams contributed equally as part of the DIEKB consortium. The Airbus team is responsible for the deep learning analysis and the development of \fake. IJN and Mondeca designed the \vago algorithm and jointly developed the lexical database and program corrections: IJN proposed the typology of vagueness and scoring rules, Mondeca implemented the \vago tool and interface. 
 The three teams contributed to the analysis and discussion of the results.
 PG, BI, GG, GA, and PE wrote the paper, which all authors read and discussed. PG, BI, GG and PE are the main authors, PG and BI are joint first. Correspondence: paul.egre@ens.fr, paul.guelorget@airbus.com, benjamin.icard@ens.fr.}

\bibliographystyle{ieeetr}
\bibliography{bibliography}

\end{document}

%% file: texts/1intro.tex
The computational verification of textual claims is motivated by the  proliferation of misinformation on the Web and online resources. Such methods are needed because human verification does not scale up to the diffusion speed of online misinformation, which spreads up to six times faster than real news~\cite{Vosoughi1146}. It is also worth mentioning that up to 50\% of the shares of viral claims happens within minutes of their appearance~\cite{DBLP:journals/corr/abs-1304-6777}.

 Extant proposals for detecting fake news have been focused on verifying textual claims given reference information, based on corpora of previously checked claims~\cite{TchechmedjievFB19,ShaarBMN20}. Some approaches exploit the thousands of reports written by journalists in fact-checking organizations to automatically verify if a claim is true or not. Other works use textual documents as trusted information to be analyzed in order to validate claims. Recently, there have been new methods to verify claims using reference-structured data, for example by adapting transformers to let them model datasets~\cite{ChenWCZWLZW20,tapas20}.

In this paper, we propose to
treat lexical vagueness as a cue for the automatic identification of fake news conveying biased information. The motivation is twofold: vague claims are less sensitive to factual refutation than precise claims \cite{Egre&Icard2018}; moreover vague terms are more prone to subjective interpretation than precise terms, particularly in the adjectival domain \cite{Kaiser&Wang2021}.
On the technical side, we use and compare two independent detection methods for identifying fake news: the first, deployed in the algorithm \vago, detects and measures vagueness and subjectivity in texts using a semantic-based approach combined with NLP techniques, while the second, deployed in the classifier \fake, relies on Convolutional Neural Network to classify texts as biased or legitimate. The motivation behind this comparison is to investigate whether \vago's and \fake's respective classification results actually converge. Our aim is thus to confront and to relate a semantic method with a deep learning method and to pool their results.

The paper is organized as follows. In section \ref{sec:vago}, we present the expert-based approach of \vago. We first motivate our emphasis on vagueness by explaining the way in which some vague expressions convey subjectivity. We present a typology of vague terms used by \vago to classify corpora as being either \textit{vague} or \textit{precise}, \textit{opinion} or \textit{factual}, as well as simple scoring rules used to implement an online text analyzer.

Section \ref{sec:deep} describes the deep learning approach of \fake. This classifier based on a Convolutional Neural Network (CNN) architecture was trained on more than 36,000 texts manually labelled as ``fake news'' or ``legitimate''. \fake exploits Class Activation Maps (CAMs) to define the contribution of each sub-part of the document to each considered class, allowing for word-level interpretation. 

In section \ref{sec:correlation}, we compare the categorization by \vago between \textit{vague} or \textit{precise}, \textit{opinion} or \textit{factual}, with the classification of news articles by \fake between \textit{legitimate} or \textit{biased}. 
In terms of explainable AI, we expect that doing so will help us see more clearly into the determinants of the \fake classification.

Finally, in Section \ref{sec:discussion} we discuss the opportunities and gains in terms of explainability of mixing expert-based systems with deep learning, data-driven classifiers. 

%% file: texts/2vago.tex
\subsection{Vagueness and subjectivity}

Vague words are expressions whose meaning is indeterminate and compatible with an open-ended range of possible interpretations\cite{russell1923vagueness}. Typical examples of vague words include gradable adjectives like ``tall'', ``rich'', ``intelligent'', whose extension is left open by the speaker \cite{kennedy2007vagueness}. When hearing ``John is tall'', the listener generally cannot infer a precise representation of John's height, but only a set of more or less probable values \cite{lassiter2017adjectival}. This is different if we hear the precise sentence ``John is exactly 187cm tall'', which eliminates more possibilities. 

There is no direct relation between a sentence being vague and its classification as true or false
\cite{Egre&Icard2018}. An utterance with precise truth conditions can perfectly be false (``the Eiffel Tower is 96 meters high''). Conversely, a vague sentence can be judged uncontroversially true (``the Eiffel Tower is a tall building''). As noted by \cite{russell1923vagueness}, however, a vague sentence has higher chances of being judged true than a precise one, because it is compatible with more possible states of affair. Relatedly, vague expressions are generally seen as more subjective than precise expressions, because both speaker and listener can interpret them in different ways \cite{verheyen2018subjectivity}. Thus, a large class of gradable adjectives is described as subjective, or even as evaluative \cite{kennedy2013two,mcnally2017aesthetic,Solt2018,Kaiser&Wang2021}.

In principle, a seasoned liar could use only precise language to create fake news. But it may be harder to ensure coherence, and it may thereby add up to the cognitive cost of having to make up a story  (see \cite{verschuere2018taxing}). In contrast, using vague language can be a cheap way of making one's utterances easier to accept, because less sensitive to factual refutation \cite{Egre&Icard2018}. Importantly, vague language is also used cooperatively to minimize error and to communicate uncertainty \cite{deemter2009utility,Egre&Icard2018, egre2020optimality}. But an overwhelming reliance on vague language may signal that the discourse is possibly less factual, and more prone to \emph{bullshitting} (see \cite{Frankfurt2009}) or to the spreading of \textit{biased} information.

\subsection{A typology of vague expressions}

To detect vagueness and subjectivity in texts, IJN and Mondeca have developed a lexical database and associated algorithm called \texttt{VAGO}.
The inventory relies
on the typology proposed in \cite{Egre&Icard2018}, which distinguishes four types of lexical vagueness: approximation ($V_A$), generality ($V_G$), degree-vagueness ($V_D$), and combinatorial vagueness ($V_C$). 

Expressions of approximation here include mostly modifiers such as ``around'', ``about'', ``almost'', ``nearly'', ``roughly'', which loosen up the meaning of the expression they modify (compare ``around 10 o'clock" and ``10 o'clock"). Expressions of generality include determiners such as ``some'' and modifiers like ``at most'', ``at least''. Unlike the former, these expressions have precise truth-conditions but they fail to give a maximally informative answer to the question ``how many'' (compare ``some students/at least three students left'' to ``eleven students left''). 

The class of degree-vague and combinatorially vague expressions includes mostly one-dimensional adjectives on the one hand (like ``tall'', ``old'', ``large'') and multi-dimensional adjectives on the other (``beautiful'', ``intelligent'', ``good'', ``qualified''), which combine several dimensions or respects of comparison. The opposition between degree-vagueness and combinatorial vagueness is adapted from \cite{alston1964philosophy}. 
In Alston's approach, combinatorial vagueness concerns not just adjectives but also common nouns and verbs. The class of degree-vague and combinatorially vague expressions consists mostly of adjectives in \vago (see Section \ref{sec:correlation}), knowing that in the case of adjectives, the \emph{degree} vs \emph{combinatorial} distinction is congruent with the distinction proposed by Kaiser and Wang between \emph{simple-subjective} vs \emph{complex-subjective} adjectives \cite{Kaiser&Wang2021}.

Thus, we assume that subjectivity is introduced foremost by expressions of type $V_D$ and $V_C$. Two competent speakers can have a non-factual disagreement as to whether someone is ``tall'' or ``old'', regardless of the existence of common scales of physical measurement, and even as they share the same knowledge of height or age \cite{kennedy2013two}. And the more dimensions attached to an expression, the more disagreement is expected to arise between competent speakers, as evidenced in the case of evaluative adjectives like ``beautiful'' or ``smart'' for which no standard scale of measurement is available \cite{mcnally2017aesthetic}. By contrast, we assume that expressions of type $V_A$ and $V_G$ give rise to no subjectivity, or at least to limited subjectivity, since the interpretation of these expressions is number-relative and less context-sensitive (``roughly 20'' is relative to the precise value 20, unlike ``tall'', and similarly ``some students'' literally means ``more than 0 students'').

As a result, expressions of type $V_A$ and $V_G$ are treated as factual expressions, and expressions of type $V_D$ and $V_C$ as subjective expressions. This means that logically the class of vague expressions is not coextensional with the class of subjective expressions, but forms a superset.



\subsection{VAGO: detection and scoring}


In \vago, vagueness and subjectivity are detected and scored bottom-up, from words to sentences to larger texts. 
For a given sentence, its vagueness score  is defined simply as the ratio of vague words to the total number of words of the sentence.

\begin{equation}
R_{vague}(\phi) =
\frac{(\overbrace{|V_G|_{\phi} + |V_A|_{\phi}}^{factual} + \overbrace{|V_D|_{\phi} + |V_C|_{\phi}}^{subjective})}{N_{\phi}}
\end{equation}

\noindent where $N_{\phi}$ designates the total number of words in the sentence $\phi$ and $|V_G|_{\phi}$, $|V_A|_{\phi}$, $|V_D|_{\phi}$ and $|V_C|_{\phi}$ denote the number of terms of each of the four types of vagueness (generality, approximation, degree-vagueness and combinatorial vagueness).

Similarly, the subjectivity score of a sentence is calculated as the ratio of subjective expressions to the total number of words in the sentence.

\begin{equation}
R_{subjective}(\phi) =
\frac{|V_D|_{\phi} + |V_C|_{\phi}}{N_{\phi}}
\end{equation}

\noindent Both the vagueness score and the subjectivity score of a sentence vary between 0 and 1. When a sentence contains at least one vague term, the degree of {vagueness} of the sentence is nonzero, $R_{vague}(\phi) > 0$. When $R_{vague}(\phi)=0$ this implies the sentence does not contain vague vocabulary, and similarly for subjectivity.

For sets of sentences, the vagueness score and the subjectivity score can be defined as the proportion of sentences with nonzero vagueness and nonzero subjectivity scores respectively. More fine-grained measures can be proposed, but these scores suffice to characterize the prevalence of each feature. If $T$ is a text and $N_T$ its total number of sentences, then:

\begin{equation}\label{eq:prop_vague}
R_{vague}(T)=\frac{|\{\phi \in T | R_{vague}(\phi)>0\}|}{N_T}
\end{equation}

\begin{equation}\label{eq:prop_subj}
R_{subjective}(T)=\frac{|\{\phi \in T | R_{subjective}(\phi)>0\}|}{N_T}
\end{equation}

\subsection{VAGO Implementation}
\texttt{VAGO}'s back-end is built on top of the popular GATE~\cite{gatearch} framework for NLP content processing. It also leverages the semantic content annotator CA-Manager~\cite{cam2013}, a semantic-based tool for knowledge extraction from unstructured data. The current pipeline automatically detects the language of the corpus (English or French) using TexCat.\footnote{\url{https://www.let.rug.nl/vannoord/TextCat/index.html}} 

The back-end uses a workflow composed of different analysis engines, combining NLP processing from GATE and semantic technologies with a UIMA-based infrastructure.\footnote{Unstructured Information Management Architecture (\url{http://uima.apache.org)}} UIMA architecture allows enriching and customizing engines to address specific needs in the application. \texttt{VAGO} is also available through HTTP REST API. 

\subsection{Online tool}

The online tool \texttt{VAGO}, available from Mondeca's website,\footnote{\url{https://research.mondeca.com/demo/vago/}} provides a graphical interface for the representation of these scores using two barometers. One barometer represents the degree to which a text is vague or precise. The other barometer indicates the degree to which the text reports an opinion or is factual (the proportion of subjective vs objective vocabulary). A section of detailed results explains for each sentence the vague triggers detected and the corresponding category ($V_X$). 

The online tool uses an anonymous profile, with some restrictions to call the service such as a limit number of characters as input (e.g., 1200), and some predefined short sample texts. The online application automatically detects the language of the text entered by the user (French or English).

\begin{figure}[H]
\captionsetup[subfigure]{labelformat=empty}
   \includegraphics[width=0.49\textwidth]{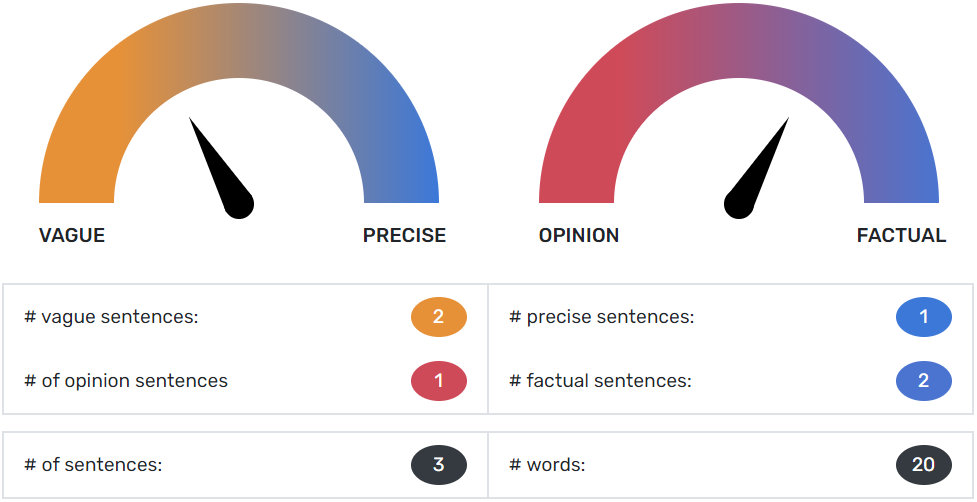}
    \caption{Results summary for text with three sentences}
    \label{fig:barometers}
\end{figure}

As a toy example, consider the text $T=\{$``\emph{Most sensational news articles are sometimes hard to believe'', ``Two plus two equals four'', ``Mary left Paris around 2pm''}$\}$, comprised of three sentences. $T$ contains five vague terms (``most'', ``sensational'', ``sometimes'', ``hard", ``around''), with ``sometimes'' and ``most'' instantiating generality, ``sensational'' and ``hard" pertaining to combinatorial vagueness, and ``around'' to approximation. While the sentence ``\emph{Two plus two equals four''} is precise, the other two are vague: \emph{``Mary left Paris around 2pm''} is only factual but ``\emph{Most sensational news articles are sometimes hard to believe''} contains both factual and subjective terms. The barometers reporting the proportion of vague and subjective sentences in the text $T$ are shown in Figure \ref{fig:barometers}.   

%% file: texts/3deep.tex
Another way to detect fake news is to learn how to distinguish them inductively, by means of examples. In this section we detail a previously developed classifier \cite{gadek2020interpretable}, \fake, and the datasets used for training.
    
\subsection{Deep learning text classification: Related work }
    
    Part of machine learning, text classification is known through numerous applications: sentiment analysis \cite{abbasi_sentiment_2008}, language identification \cite{castro_smoothed_2017} or even emotion detection \cite{liu2019oasis}, dealing with a variety of texts from tweets to Wikipedia articles.
    While very efficient, these methods often lack explainability as most of them work in a ``black box'' mode.
    
Explaining a prediction on a text means to characterize sub-document segments: this can be performed either by observing the behavior of an attention mechanism \cite{bahdanau_neural_2014}; another way relies on the extraction of class activation maps \cite{zhou_learning_2016}. 


Regarding the first approaches, a self-attention mechanism was introduced in \cite{lin_structured_2017} for text classification. 
Attention scores are computed as the salience of a sub-part of the document relatively to a generic classification task, and not relatively to a precise label. Indeed this mechanism focuses on the ``important'' words, not on the words that justify an isolated prediction. Ranging from 0 (no interest) to 1 (high interest), the attention scores do not carry information about the prediction. However, high-attention words and phrases act with more impact, either for or against class membership.

The second approach, \textit{class activation maps} (CAMs), was introduced in \cite{zhou_learning_2016}. CAMs are literally the contribution of each sub-part of the document to each considered class. 
In order to preserve a direct link with the input, CAMs require specific architecture conditions detailed in the next paragraph. 

\subsection{CAM to explain text classification}
\label{sec:cam}
CAMs were first introduced in image classification \cite{zhou_learning_2016}, and only recently adapted to text classification \cite{guelorget_deep_2020}. The intuition relies on the fact that spatiality is preserved across convolutional layers, whereas it is lost in the last fully-connected layers used in most classifiers. Thus this method concerns fully-convolutional networks; the final prediction layer can rely either on global average pooling (GAP) or on global max pooling (GMP), which is applied to squash the $T$ spatial features vectors associated to the deepest feature maps $F = \{ F_1, ..., F_T \} \in \mathbb{R}^{T \times K}$ into a single, global feature vector $F_g\in\mathbb{R}^K$ in which all spatiality is lost. $K$ denotes the size of the considered feature maps. 

The technical details of the \fake model implementation are the following: it consists of a fully-convolutional neural network made of 3 layers of 128 kernels of size 5. The final layer consists of a global average pooling and a softmax classification. As inputs, we rely on a pre-trained word embedding, of dimension 300: FastText \cite{joulin_bag_2016}. Indeed, a more recent embedding, even if context-aware such as BERT \cite{devlin_bert:_2018} or Elmo \cite{peters_deep_2018}, would mean a much higher-complexity architecture, detrimental to the interpretability of the class prediction (as it would take into account the role of the neighboring words without explaining how). The texts are only tokenized (words and punctuation items) through the Python library NLTK\footnote{\url{https://www.nltk.org/}}. These pre-trained embeddings imply that the model has to grasp the context: \fake does it thanks to its convolutional layers. Note that \fake traces the importance of each token's contribution.

Because CAMs require to preserve the temporality across layers, \fake uses the same padding for all convolutions to have exactly one output layer directly corresponding to each input token. As a result the last convolutional layer presents a number of outputs that is equal to the number of input words: the $t^{th}$ output of the last convolutional layer describes the $t^{th}$ word of the input sentence, considering also its context within the convolutional receptive field. 

The CAM extraction is explained hereafter. If there are $C$ different labels, then the softmax input is defined as:

\begin{equation}
S = W^TF_g + b
\end{equation}

\noindent where $W = \{w_c^k\} \in \mathbb{R}^{K\times C}$ and $b\in \mathbb{R}^C$ are the final weights and biases.
For any input example $x$, the class activation map for label $c$ at location $t$ is obtained by summing the contribution $w_c^kF_t^k$ of each scalar feature $F_t^k$ to the final score of label $c$, as described in the following equation:

\begin{equation}
\label{eq:CAM-definition}
\text{CAM}(x,c,t) = \sum_k{w_c^k F_t^k}
\end{equation}

\begin{figure}[htbp]
	\centering
	\includegraphics[width=0.49\textwidth]{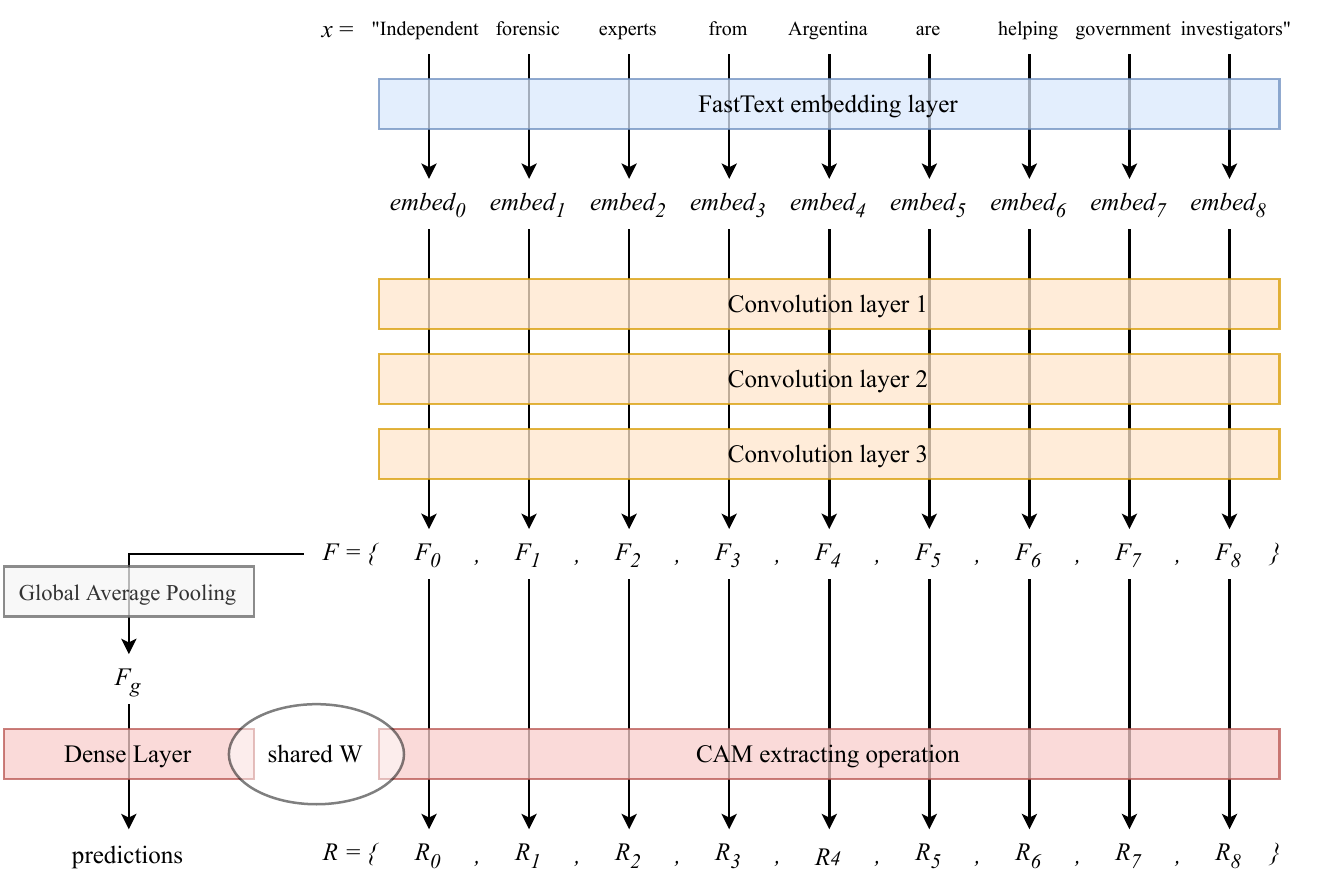}
	\caption{Extraction of Class Activation Maps from a tokenized sentence, from \cite{gadek2020interpretable}.}
	\label{fig:schema2}
\end{figure}

As a result, in a text classification context, CAMs provide a signed contribution score of each word (or punctuation item) to each class membership tackled by the model. 
Our CAM-extracted model is illustrated in Figure~\ref{fig:schema2}.

\subsection{Predicting fake-like news articles}
We trained this CAMs-based architecture to detect ``fake news'', using the following well-known press articles datasets:

\begin{itemize}
    \item Kaggle fake news\footnote{\url{https://www.kaggle.com/mrisdal/fake-news}} is composed of 13,000 documents in English, tagged as ``fake'': either ``\textit{biased}'' or ``\textit{bullshit}''. We decided to merge these two ``fake'' categories under a single ``\textit{biased}'' label as the first iterations of this classifier did not distinguish well these two kinds of bad-quality content. Moreover, doing so enables the aggregation with the next datasets.
    
    \item ISOT\footnote{\url{https://www.uvic.ca/engineering/ece/isot/datasets/fake-news/index.php}} contains 21,417 ``true'' and 23,481 ``fake'' articles, subsequently referred to as ISOT-True and ISOT-Fake.
    
    \item SignalMedia1M\footnote{\url{http://research.signalmedia.co/newsir16/signal-dataset.html}} gathers a huge corpus of press articles (in English), deemed to be legitimate \cite{katsaros2019machine}. From this resource, we select two random samples:
    \begin{itemize}
        \item 15,064 articles to balance the training,
        \item 20,071 other articles for the evaluation part.
    \end{itemize}
    
\end{itemize}
On Kaggle and ISOT, a random train/test split policy is applied (80\%/20\%). The resulting datasets are then aggregated with their SignalMedia1M complement. 


A recurrent objection on fake news classification is the risk to overfit the classifier on a topic-specific dataset, over-focusing on a few words. Particularly hard to analyze, this problem stems from the scarcity of similarly-labeled, thematically different datasets. \fake presents an advantage in order to investigate this point, through the class activation mappings: high-score tokens are easily identified.

\subsection{Validating the classifier on unseen data}

\begin{figure}[htbp]
    \centering
    \includegraphics[width=0.49\textwidth]{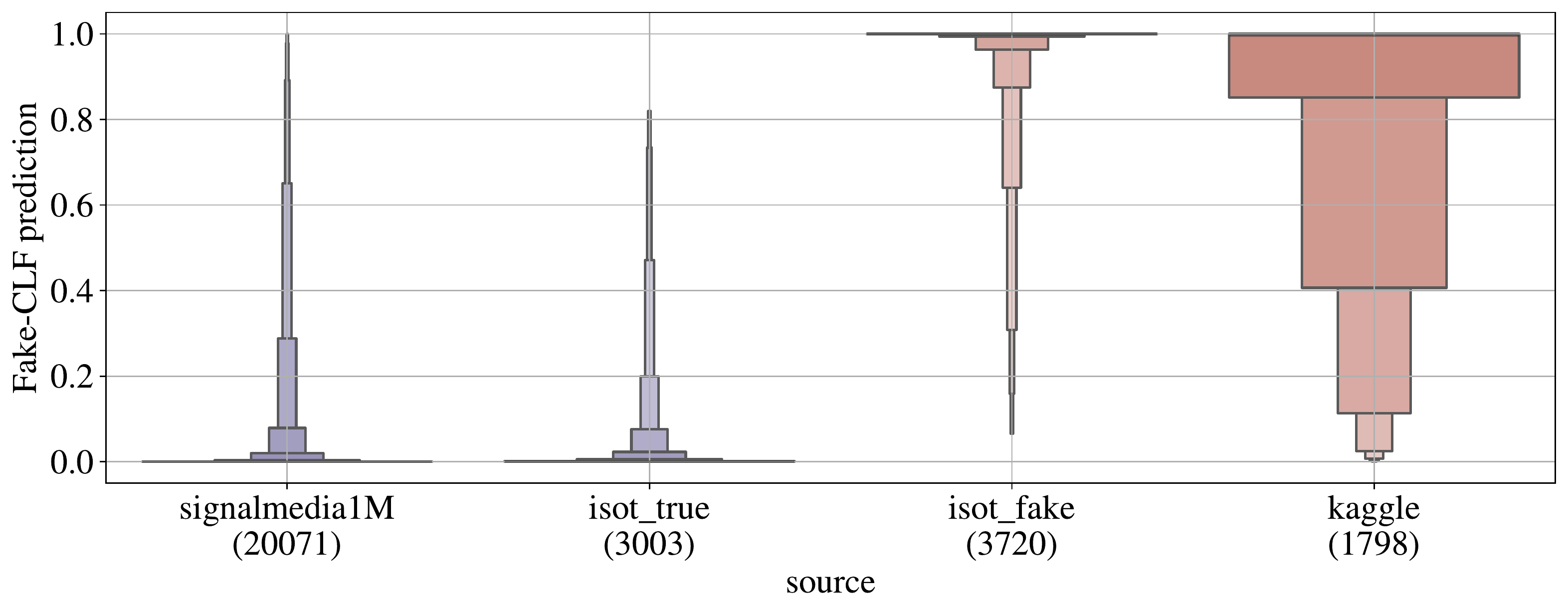}
    \caption{Classification of source corpora in the \textit{test} dataset as \textit{legitimate} (0) or \textit{biased} (1) by \fake.
    }
    \label{fig:clf_results_per_source}
\end{figure}

The training dataset of \fake consists of the four corpora previously listed (ISOT-True, ISOT-Fake, SignalMedia1M, Kaggle). The measure of the model performance with respect to the four corpora is presented on Figure \ref{fig:clf_results_per_source}. The y-axis is the output score; 0 means \textit{legitimate} while 1 stands for a \textit{biased} content. The x-axis shows the source corpora. The letter-value plots\footnote{The widest box stretches between the first and last quartiles, the second widest boxes stretch between the first/last quartile and the first/last octile, the third widest between the first/last octile and the first/last hexadeciles, and so forth \cite{letter-value-plot}.} illustrate the distribution of articles that were part of the test dataset (20\% of total data, excluded from the training set).

ISOT-True and ISOT-Fake are two facets of the ISOT dataset. ``True'' news, labeled as \textit{legitimate} articles, and ``fake'' articles, labeled as \textit{biased}, are very well recognized and separated. The same trend is also remarkable in separating SignalMedia1M and Kaggle's datasets, even though the latter shows a fuzzier distribution. Overall the classification performance on these {test} datasets results in an F1-score $= 0.955$, indicative of high accuracy. 

%% file: texts/4comparison.tex

We now turn to the comparison between the classification obtained by \fake and the measures produced by \vago.

\subsection{Experimental Settings}

The lexicon used by \vago at the time of this experiment consisted of $1,527$ English terms, and $1,150$ French terms (version of the database of 12/22/2020)\cite{datasetvago2021}. The English lexicon is comprised 95\% of adjectives, all in the classes $V_D$ and $V_C$, the whole vocabulary being distributed as follows: $|V_A|= 8$, $|V_G|= 14$, $|V_D|= 35$ and $|V_C|= 1,470$. Hence 96\% of the English lexicon consists of $V_C$ items. The French lexicon is comprised 97\% of adjectives, again all in the categories $V_D$ and $V_C$. In French, we have $|V_A|= 7$, $|V_G|= 15$, $|V_D|= 28$ and $|V_C|= 1,100$. Accordingly, 96\% of the French lexicon consists of $V_C$  vocabulary. Although the corpora used contained mostly English sentences, they included some French excerpts, which were included for analysis. Hence part of the French lexicon was used.

\subsection{Comparison and correlation of \vago and \fake}


We measure the relation between vagueness and how much articles are predicted to be manipulative by \fake by looking at two classes of vague items detected by \vago, including: all vague items ($V_A + V_G + V_D + V_C$), and subjective vague items ($V_D + V_C$). 

\begin{figure}[htbp]
    \footnotesize
    \begin{tabular}{cc}
        a. & \!\!\!\!\!\!\!\! \includegraphics[align=c,width=0.94\columnwidth]{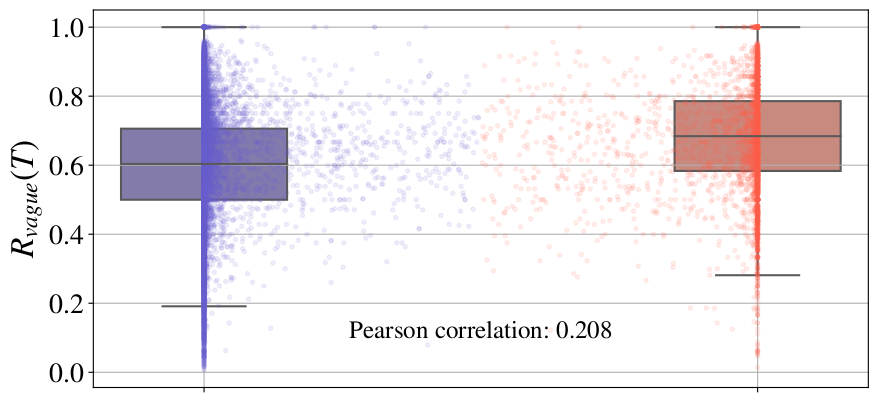} \\
        b. & \!\!\!\!\!\!\!\! \includegraphics[align=c,width=0.94\columnwidth]{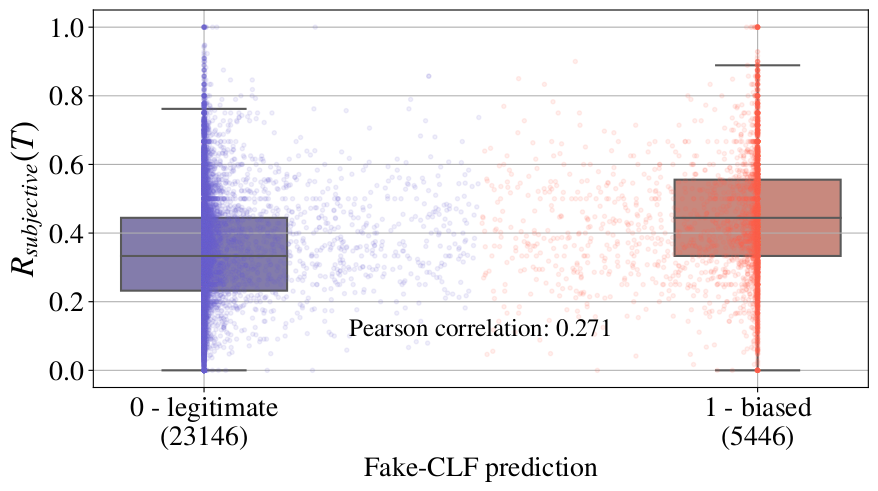}
    \end{tabular}
    
    \caption{Comparison between \fake predictions and (a) ratio of vague sentences, (b) ratio of subjective sentences
    in texts, as predicted by \vago.}
    \label{fig:vago_sentences_ratio}
\end{figure}

The relation between articles predicted to be manipulative and the presence of vagueness is measured in Figure \ref{fig:vago_sentences_ratio}~a), using \vago as our detector of vague sentences, and the measure reported in Eq. (\ref{eq:prop_vague}) for vagueness. On the left hand side, the x-axis shows the predicted class (\textit{legitimate} in blue, \textit{biased} in red) by \fake. The y-axis displays the percentage of sentences containing markers of vagueness (the higher, the more vague). The Pearson correlation coefficient between these two variables (\textit{biased}, and \textit{vague}) is 0.208 (sample size $=28,692$): \textit{biased} articles tend to contain more vague sentences, but vagueness is not the sole determining factor.








A similar analysis is performed using VAGO as a detector of \textit{subjective} sentences, also described as opinion sentences, using the measure of subjectivity reported in Eq. (\ref{eq:prop_subj}); this analysis is displayed in Figure \ref{fig:vago_sentences_ratio}~b). There, the y-axis displays the ratio of sentences that contain $V_C$ and $V_D$ type keywords. Here the correlation between these two variables is 0.271: articles detected as \textit{biased} tend to contain more markers of subjectivity.





\begin{figure}[htbp]
    \scriptsize
    \begin{tabular}{cm{.05\columnwidth}cm{.9\columnwidth}}
        a. & \includegraphics[width=0.9\columnwidth]{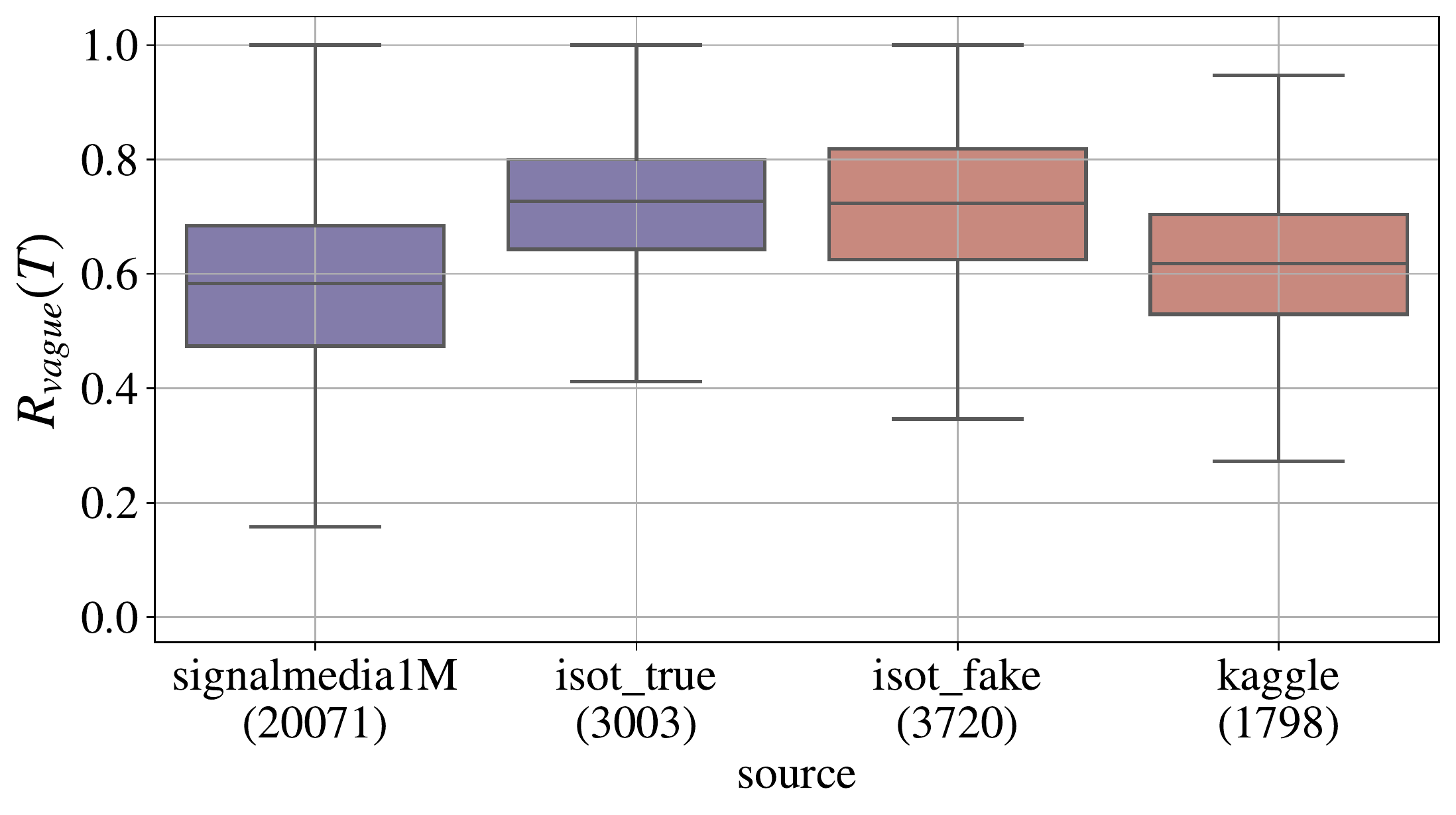} \\
        b. & \includegraphics[width=0.9\columnwidth]{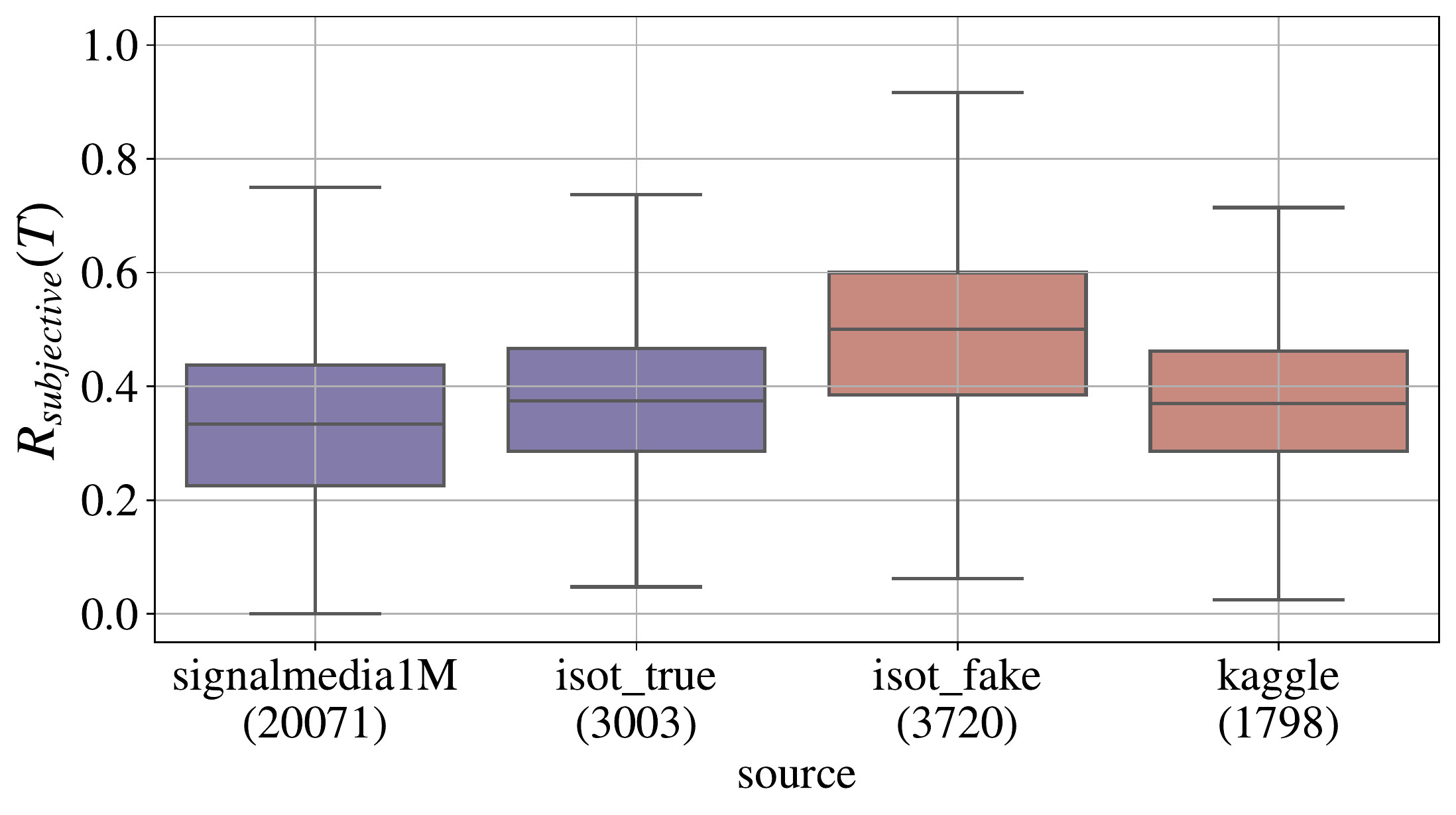} \\
    \end{tabular}
    
    \caption{Distribution of vague sentences ratio (a), vague-subjective sentences ratio (b) 
    across sources.}
    \label{fig:distrib_vague_sources}
\end{figure}

A split according to the source corpora provides further insights. Regarding subjectivity, Figure \ref{fig:distrib_vague_sources}~b) shows that corpora follow the tendency of their \textit{legitimate} vs. \textit{biased} affiliation: ISOT-true and SignalMedia1M texts are detected as less subjective than ISOT-fake and Kaggle texts. While ISOT-true documents and ISOT-fake documents contain the same proportion of vague terms all types included (Figure \ref{fig:distrib_vague_sources}~a), the difference between them concerns the occurrence of subjective vocabulary (Figure \ref{fig:distrib_vague_sources}~b). 




\subsection{Word-level analysis: exploiting the deep to find new keywords}

\begin{figure}[htbp]
    \centering
    \includegraphics[width=0.49\textwidth]{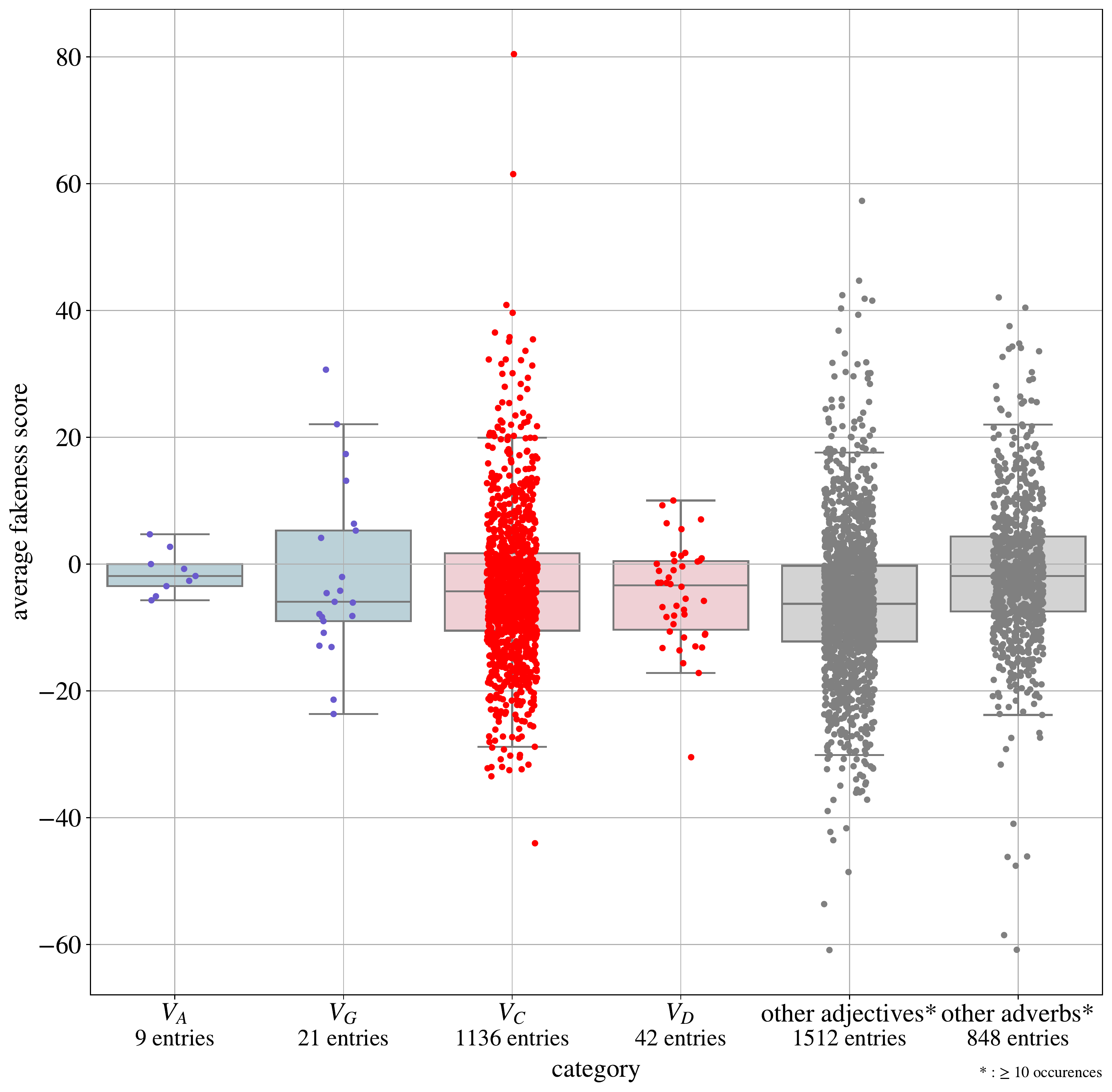}
    \caption{Average ``fakeness'' scores of \vago entries and other adjectives and adverbs according to \fake class attention maps, defined in Equation~\ref{eq:CAM-definition}.}
    \label{fig:category_distrib}
\end{figure}

\fake not only provides a prediction at article-level, but also a word-level contextual score to enable an understanding of its predictions. In Figure \ref{fig:category_distrib} we display the distribution of the average ``fakeness'' score (the CAM word-level score) of the occurring \vago keywords, as they are perceived by the classifier.
Each dot represents the average ``fakeness'' score of a vocabulary entry across all its occurrences. For each category, a boxplot sums up the score distribution.
$V_A$ and $V_G$ words used to identify \textit{vague-factual} sentences are in blue; $V_C$ and $V_D$ words used to identify \textit{vague-subjective} sentences are in red. Gray bars give the scores for other adjectives and adverbs, as a reference, as \vago contains mostly adjectives. Non-\vago entries have been tagged as adjectives or adverbs by jointly using TextBlob\footnote{\url{https://github.com/sloria/TextBlob}} and NLTK part-of-speech tagging.

As described in section \ref{sec:cam}, for every token in a sentence the CAM provides a signed score accounting for the contribution of this token and its near context to the prediction of the whole sentence. Thus, we consider that a high CAM score stands for a highly bias-inducing word according to \fake. We can see from this figure that belonging to a vague category is not sufficient to trigger a \textit{biased} or \textit{legitimate} label; however some words remain good cues in that respect. In Table \ref{tab:fakest_adjectives}, we propose a list of the thirty most bias-inducing adjectives and adverbs (considering the prevalence of adjectives in \vago and the proximity of adverbs to adjectives). The first two columns present the number of occurrences of each term and its average ``fakeness'' score. The third gives the semantic type (when listed in \vago) and the fourth the syntactic category.

\begin{table}[htbp]
    \centering
    \begin{tabular}{c|ccccc}
            word & occ & avg & \vago & part-of-speech \\ \hline
            
        \textbf{sociable} & 32 & 80.42 & V\textsubscript{C} & adj.\\
        \textbf{disgusting} & 231 & 61.50 & V\textsubscript{C} & adj.\\
        \color{gray!50}stumble & \color{gray!50}28 & \color{gray!50}57.28 & \color{gray!50} & \color{gray!50}adj.\\
        Orwellian & 13 & 44.69 &  & adj.\\
        sociopathic & 12 & 42.40 &  & adj.\\
        arrogantly & 13 & 42.04 &  & adv.\\
        misogynistic & 37 & 41.84 &  & adj.\\
        entire & 2246 & 41.54 &  & adj.\\
        \textbf{idiotic} & 48 & 40.86 & V\textsubscript{C} & adj.\\
        courageously & 15 & 40.43 &  & adv.\\
        neoconservative & 25 & 40.30 &  & adj.\\
        neoliberal & 45 & 39.31 &  & adj.\\
        coincidentally & 24 & 37.52 &  & adv.\\
        subliminal & 14 & 36.81 &  & adj.\\
        \textbf{delusional} & 61 & 35.78 & V\textsubscript{C} & adj.\\
        \textbf{pitiful} & 18 & 35.43 & V\textsubscript{C} & adj.\\
        frighteningly & 10 & 34.78 &  & adv.\\
        disturbingly & 16 & 34.30 &  & adv.\\
        shamelessly & 26 & 34.08 &  & adv.\\
        blatantly & 75 & 33.92 &  & adv.\\
        \textbf{astonishing} & 100 & 33.62 & V\textsubscript{C} & adj.\\
        laughably & 11 & 33.56 &  & adv.\\
        \color{gray!50}sic & \color{gray!50}101 & \color{gray!50}33.21 & \color{gray!50} & \color{gray!50}adj.\\
        outrageously & 18 & 32.67 &  & adv.\\
        \textbf{racist} & 971 & 32.27 & V\textsubscript{C} & adj.\\
        \textbf{massive} & 1314 & 32.14 & V\textsubscript{C} & adj.\\
        devious & 13 & 31.81 &  & adj.\\
        transnational & 52 & 31.73 &  & adj.\\
        \textbf{deplorable} & 85 & 31.56 & V\textsubscript{C} & adj.\\
        Siberian & 11 & 31.51 &  & adj.\\
        
    \end{tabular}
    \caption{Thirty most bias-inducing terms according to \fake, filtered to adjectives and adverbs with at least 10 occurrences. Entries are sorted by descending average CAM scores (avg column). \vago terms are in bold. Miscategorized adjectives and adverbs are in gray. 
    }
    \label{tab:fakest_adjectives}
\end{table}

\begin{table}[htbp]
    \centering
    \begin{tabular}{c|ccccc}
            word & occ & avg & \vago & part-of-speech \\ \hline

        \color{gray!50}Carly & \color{gray!50}16 & \color{gray!50}-60.89 & \color{gray!50} & \color{gray!50}adj.\\
        \color{gray!50}Emily & \color{gray!50}72 & \color{gray!50}-60.84 & \color{gray!50} & \color{gray!50}adv.\\
        provisionally & 28 & -58.53 &  & adv.\\
        40th & 32 & -53.64 &  & adj.\\
        \color{gray!50}Experian & \color{gray!50}40 & \color{gray!50}-48.58 & \color{gray!50} & \color{gray!50}adj.\\
        \color{gray!50}vSphere & \color{gray!50}14 & \color{gray!50}-47.59 & \color{gray!50} & \color{gray!50}adv.\\
        upwardly & 11 & -46.20 &  & adv.\\
        fortnightly & 11 & -46.13 &  & adv.\\
        \textbf{cloudy} & 99 & -44.04 & V\textsubscript{C} & adj.\\
        sectoral & 14 & -43.56 &  & adj.\\
        unbeaten & 113 & -42.26 &  & adj.\\
        \color{gray!50}trimble & \color{gray!50}19 & \color{gray!50}-41.68 & \color{gray!50} & \color{gray!50}adj.\\
        \color{gray!50}premiere & \color{gray!50}57 & \color{gray!50}-40.97 & \color{gray!50} & \color{gray!50}adv.\\
        \color{gray!50}Sebastian & \color{gray!50}80 & \color{gray!50}-38.96 & \color{gray!50} & \color{gray!50}adj.\\
        directorial & 12 & -37.21 &  & adj.\\
        playable & 18 & -37.18 &  & adj.\\
        Tanzanian & 11 & -36.07 &  & adj.\\
        treble & 19 & -36.01 &  & adj.\\
        undertaken & 25 & -35.82 &  & adj.\\
        interactive & 378 & -35.57 &  & adj.\\
        \color{gray!50}semifinal & \color{gray!50}19 & \color{gray!50}-34.96 & \color{gray!50}  & \color{gray!50}adj.\\
        topographic & 16 & -34.77 &  & adj.\\
        shareable & 10 & -34.44 &  & adj.\\
        \color{gray!50}Dorian & \color{gray!50}10 & \color{gray!50}-33.97 & \color{gray!50} & \color{gray!50}adj.\\
        procedural & 92 & -33.50 &  & adj.\\
        \textbf{muddy} & 46 & -33.48 & V\textsubscript{C} & adj.\\
        unrivalled & 13 & -33.24 &  & adj.\\
        \textbf{comprehensive} & 1001 & -32.52 & V\textsubscript{C} & adj.\\
        \textbf{moody} & 146 & -32.37 & V\textsubscript{C} & adj.\\
        unauthorised & 14 & -32.36 &  & adj.\\
    \end{tabular}
    \caption{Thirty least bias-inducing terms  according to \fake, filtered to adjectives and adverbs with at least 10 occurrences. Entries are sorted by ascending average CAM scores (avg column). \vago terms are in bold. Miscategorized adjectives and adverbs are in gray.
    }
    \label{tab:legitest_adjectives}
\end{table}

  Even though two part-of-speech tagging tools were used to identify ``other  adjectives''  and  ``other adverbs'', some artifacts (displayed in gray) are wrongly tagged as adjectives or adverbs. Several evaluative adjectives like
 ``disgusting'' and ``racist'' are identified by \vago as markers of subjectivity; others like ``Orwellian'', ``sociopathic'' and ``arrogantly'' are not in the \vago database but are also clear vehicles of subjectivity and evaluativity. In fact, disregarding the two artifacts marked in gray, all terms in Table \ref{tab:fakest_adjectives} are adjectives or adverbs that fall under the category of combinatorial vagueness and that could be added as such to the \vago lexicon.
 
 
A comparison with least bias-inducing terms (scores ranging $[-60; -30]$, Table \ref{tab:legitest_adjectives}) shows a more heterogeneous set. Fewer terms are already part of the \vago lexicon. Furthermore, among adjectives and adverbs, several are non-gradable and precise (``40th'', ``unbeaten'', ``upwardly'', ``fortnightly'', ``Tanzanian'', ``unauthorized''). Fewer correspond to evaluative adjectives or to adjectives that would be eligible for the type $V_C$ (``interactive'', ``unrivalled'' may be exceptions).

  



%% file: texts/5discussion.tex


Our hypothesis in this paper was that because vague terms are often subjective, the prevalence of vagueness in discourse can be used as a cue that the discourse conveys fake news (whether \emph{biased} or \emph{bullshit}). In order to test this hypothesis, we pooled two distinct methodologies. The first methodology involved the semantic-based algorithm \texttt{VAGO}, which provides a measure of vagueness vs precision in text, and a measure of subjectivity vs objectivity in text. The second methodology involved the deep learning classifier \fake, relying on CNN and CAM techniques. 

By comparing the results of \fake with the measures of vagueness and subjectivity obtained by \vago on four distinct corpora pre-identified as containing ``fake'' vs ``true'' articles, we found a positive correlation between vagueness and fakeness, as well as between subjectivity and fakeness. The association with fakeness is stronger for subjective vague terms (of type $V_D$ and $V_C$ in \vago's taxonomy). While an overwhelming majority of the terms in the \vago database are markers of subjectivity, this stronger association also confirms that not all types of vagueness introduce subjectivity.


From a methodological point of view, the two approaches discussed in this paper can be viewed as complementary. In one direction, \fake identifies as markers of fakeness some adjectives not originally in the \texttt{VAGO} database, but which clearly belong to the class $V_C$ of adjectives exemplifying multi-dimensional vagueness, and conveying subjectivity. The correlation found is therefore expected to increase by the inclusion of a larger vocabulary.
In the converse direction, the typology deployed in \texttt{VAGO} helps to make the results of \fake more easily interpretable and explainable. Indeed, while the inductive generalizations operated by \fake remain opaque, \texttt{VAGO} rests on a transparent semantic architecture. More work is needed to narrow the gap between the two approaches, but the consistency in the correlation between the two classifications is a step toward increased explainability.

%% file: texts/6conclusion.tex

Focusing on lexical vagueness as a marker of subjectivity, we have combined a semantic-based NLP engine \vago with a deep learning classifier \fake to improve the detection of fake news contents. 
The two approaches yield convergent results and they each provide useful input to improve the other. 

Several points remain for further elaboration. First of all, the \vago lexicon was still limited at the time of this comparison. We have seen how it can be expanded using \fake, but the base will keep evolving. 
In particular, other expressions beside adjectives (and adverbs) can introduce vagueness and subjectivity. One non-adjective that was included in \vago at the time of this comparison is the modal ``should''. ``Should'' was tentatively classified as $V_C$ considering its expressive dimension in  deontic utterances stating moral prescriptions. As it turns out, \fake predicts that ``should'' is not among the most bias-inducing terms when more entries than adjectives and adverbs are considered for analysis. Further work is needed to adjudicate the potential vagueness and subjectivity of ``should'' and related modals.


Secondly, the correlation found between \textit{biased} predictions and subjectivity scores is limited, thus subjectivity explains only part of the variance in the classification of \fake. This is not surprising, since subjectivity is only one among several factors that can contribute to misinformation. Fake news consist not only in opinionated texts, but also in clever fabrications, hoaxes, or simply mistaken factual reports \cite{rubin2015deception}. The type of integration proposed between \fake and \vago should therefore be extended to explore other sources of fakeness beside bias.









%

%